\documentclass[runningheads]{llncs}
\usepackage{booktabs}       
\usepackage[hyphens]{url}            

\usepackage{cite} 
\usepackage{graphicx}
\usepackage{amsmath}
\usepackage{color}
\usepackage{caption}
\usepackage{subcaption}
\usepackage{pgfplots}
\usepackage{tikz}
\usepackage{cleveref}
\Crefname{equation}{Eq.}{Eqs.}
\Crefname{figure}{Fig.}{Figs.}
\Crefname{tabular}{Tab.}{Tabs.}

\usetikzlibrary{arrows.meta}
\usetikzlibrary{decorations.markings}
\DeclareUnicodeCharacter{2212}{-}

\begin{document}
%
\title{Iterative SE(3)-Transformers}
%
%
\author{Fabian B. Fuchs\,$^\dagger$\inst{1} \and
Edward Wagstaff\,$^\dagger$\inst{1} \and
Justas Dauparas\inst{2} \and
Ingmar Posner\inst{1}}
\authorrunning{F. Fuchs and E. Wagstaff et al.}
%
\institute{
Department of Engineering Science, University of Oxford, Oxford, UK \and
Institute for Protein Design, University of Washington, WA, USA
\newline \vspace{2mm} $\dagger$ These authors contributed equally
\newline \vspace{2mm} \email{\{fabian,ed\}@robots.ox.ac.uk}\
}
\maketitle              
\begin{abstract}

When manipulating three-dimensional data, it is possible to ensure that rotational and translational symmetries are respected by applying so-called \emph{SE(3)-equivariant} models.
Protein structure prediction is a prominent example of a task which displays these symmetries. Recent work in this area has successfully made use of an SE(3)-equivariant model, applying an iterative SE(3)-equivariant attention mechanism.
Motivated by this application, we implement an iterative version of the SE(3)-Transformer, an SE(3)-equivariant attention-based model for graph data.
We address the additional complications which arise when applying the SE(3)-Transformer in an iterative fashion, compare the iterative and single-pass versions on a toy problem, and consider why an iterative model may be beneficial in some problem settings.
We make the code for our implementation available to the community\footnote{\url{https://github.com/FabianFuchsML/se3-transformer-public}}.

\end{abstract}

\section{Introduction}


Tasks involving manipulation of three-dimensional (3D) data often exhibit rotational and translational symmetry, such that the overall orientation or position of the data is not relevant to solving the task. One prominent example of such a task is protein structure refinement \cite{adiyaman2019methods}. The goal is to improve on the initial 3D structure -- the position and orientation of the structure, i.e. the frame of reference, is not important to the goal. We would like to find a mapping from the initial structure to the final structure such that if the initial structure is rotated and translated then the predicted final structure is rotated and translated in the same way. This symmetry between input and output is known as \emph{equivariance}.
More specifically, the group of translations and rotations in 3D is called the special Euclidean group and is denoted by SE(3). The relevant symmetry is known as SE(3) equivariance.

In the latest Community-Wide Experiment on the Critical Assessment of Techniques for Protein Structure Prediction
(CASP14) structure-prediction challenge, DeepMind's AlphaFold 2 team \cite{jumper2020high} successfully applied machine learning techniques to win the categories ``regular targets'' and ``interdomain prediction'' by a wide margin. This is a important achievement as it opens up new routes for understanding diseases and drug discovery in cases where protein structures cannot be experimentally determined \cite{kuhlman2019advances}.
At this time, the full implementation details of the AlphaFold 2 algorithm are not public. Consequently, there is a lot of interest in understanding and reimplementing AlphaFold 2\cite{BlogCarlos,openfold2,wang2021pytorch,markowitz2020alphafold,alquraishi2020alphafold}.

One of the core aspects that enabled AlphaFold 2 to produce very high quality structures is the end-to-end iterative refinement of protein structures \cite{jumper2020high}.
Concretely, the inputs for the refinement task are the estimated coordinates of the protein, and the outputs are updates to these coordinates.
This task is equivariant: when rotating the input, the update vectors are rotated identically.
To leverage this symmetry, AlphaFold 2 uses an SE(3)-equivariant attention network.
The first such SE(3)-equivariant attention network described in the literature is the SE(3)-Transformer \cite{SE3Transformer}.
However, in the original paper, the SE(3)-Transformer is only described as a single-pass predictor, and its use in an iterative fashion is not considered.

In this paper, we present an implementation of an iterative version of the SE(3)-Transformer, with a discussion of the additional complications which arise in this iterative setting. In particular, the backward pass is altered, as the gradient of the loss with respect to the model parameters could flow through basis functions. We conduct toy experiments to compare the iterative and single-pass versions of the architecture, draw conclusions about why this architecture choice has been made in the context of protein structure prediction, and consider in which other scenarios it may be a useful choice. The code will be made publicly available$^3$.

\section{Background}

In this section we provide a brief overview of SE(3) equivariance, with a description of some prominent SE(3)-equivariant machine learning models. To situate this discussion in a concrete setting, we take motivation from the task of protein structure prediction and refinement, which is subject to SE(3) symmetry.

\subsection{Protein Structure Prediction and Refinement}

In protein structure prediction, we are given a target sequence of amino acids, and our task is to return 3D coordinates of all the atoms in the encoded protein. Additional information is often needed to solve this problem -- the target sequence may be used to find similar sequences and related structures in protein databases first \cite{uniprot2019uniprot, wwpdb2019protein}.
Such coevolutionary data can be used to predict likely interresidue distances using deep learning -- an approach that has been dominating protein structure prediction in recent years \cite{senior2020improved, xu2019distance, yang2020improved}.
Coevolutionary information is encoded in a multiple sequence alignment (MSA) \cite{steinegger2019hh}, which can be used to learn pairwise features such as residue distances and orientations \cite{yang2020improved}. These pairwise features are constraints on the structure of the protein, which inform the prediction of the output structure. One could start with random 3D coordinates for the protein chain and use constraints from learnt pairwise features to find the best structure according to those constraints. The problem can then be approached in an iterative way, by feeding the new coordinates back in as inputs and further improving the structure. This iterative approach can help to improve predictions \cite{greener2019deep}.

Importantly, MSA and pairwise features do not include a global orientation of the protein -- in other words, they are invariant under rotation of the protein. This allows for the application of SE(3)-equivariant networks, which respect this invariance by design, as done in AlphaFold 2 \cite{jumper2020high}. The predictions of an SE(3)-equivariant network, in this case predicted shifts to backbone and side chain atoms, are always relative to the arbitrary input frame of reference, without the need for data augmentation. SE(3)-equivariant networks may also be applied in an iterative fashion, and when doing so it is possible to propagate gradients through the whole structure prediction pipeline. This full gradient propagation contrasts with the disconnected structure refinement pipeline of the first version of AlphaFold \cite{senior2020improved}.

\subsection{Equivariance and the SE(3)-Transformer}
A function, task, feature\footnote{A \emph{feature} of a neural network is an input or output of any layer of the network.}, or neural network  is \emph{equivariant} if transforming the input results in an equivalent transformation of the output. Using rotations $\mathbf{R}$ as an example, this condition reads:

\begin{align}
    f(\mathbf{R} \cdot \vec{x}) = \mathbf{R} \cdot f(\vec{x}) 
\end{align}

In the following, we will focus on 3D rotations only -- this group of rotations is denoted SO(3). Adding translation equivariance, i.e. going from SO(3) to SE(3), is easily achieved by considering relative positions or by subtracting the center of mass from all coordinates.

A set of data relating to points in 3D may be represented as a graph. 
Each node has spatial coordinates, as well as an associated feature vector which encodes further relevant data.
A node could represent an atom, with a feature vector describing its momentum.
Each edge of the graph also has a feature vector, which encodes data about interactions between pairs of nodes.
In an equivariant problem, it is crucial that equivariance applies not only to the positions of the nodes, but also to all feature vectors -- for SO(3) equivariance, the feature vectors must rotate to match any rotation of the input.
To distinguish such equivariant features from ordinary neural network features, we refer to them as \emph{fibers}.

A fiber could, for example, encode momentum and mass as a 4 dimensional vector, formed by concatenating the two components. A momentum vector would typically be 3 dimensional (also called \textit{type-1}), and is rotated by a 3x3 matrix. The mass is scalar information (also called \textit{type-0}), and is invariant to rotation. The concept of \textit{types} comes from representation theory, where a type-$\ell$ feature is rotated by a $(2\ell+1) \times (2\ell+1)$ Wigner-D matrix.\footnote{We can think of type-0 features as rotating by the $1\times 1$ rotation matrix $(1)$.} Because a fiber is a concatenation of features of different types, the entire fiber is rotated by a block diagonal matrix, where each block is a Wigner-D matrix. \cite{ThomasSKYKR18,WeilerGWBC18,kondor2018nbody}

At the input and output layers, the fiber structure is determined by the task at hand. For the intermediate layers, arbitrary fiber structures can be chosen. In the following, we denote the structure of a fiber as a dictionary. E.g. a fiber with 3 scalar values (e.g. RGB colour channels) and one velocity vector has 3 type-0 and 1 type-1 feature: \texttt{\{0:3, 1:1\}}. This fiber is a feature vector of length 6.

There is a large and active literature on machine learning methods for graph data, most importantly graph neural networks \cite{KipfFWWZ18,VelikovicCCRLB2017,wang2017nonlocal,Battaglia2018relational,Wu2020comprehensive}. The SE(3)-Transformer \cite{SE3Transformer} in particular is a graph neural network explicitly designed for SE(3)-equivariant tasks, making use of the fiber structure and Wigner-D matrices discussed above to enforce equivariance of all features at every layer of the network.

\textbf{Alternative Approaches to Equivariance:} The Wigner-D matrix approach\footnote{This approach rests on the theory of \emph{irreducible representations\cite{ThomasSKYKR18,WeilerGWBC18}}.} at the core of the SE(3)-Transformer is based on closely related earlier works \cite{ThomasSKYKR18,WeilerGWBC18,kondor2018nbody}. In contrast, Cohen et al. \cite{CohenW16} introduced rotation equivariance by storing copies corresponding to each element of the group in the hidden layers -- an approach called regular representations. This was constrained to 90 degree rotations of images. Two recent regular representation approaches \cite{Finzi2020,LieTransformer} extend this to continuous data by sampling the (infinite) group elements and map the group elements to the corresponding Lie group to achieve a smoother representation.

\subsection{Equivariant Attention}

The second core aspect of an SE(3)-Transformer layer is the self-attention \cite{VaswaniSPUJGKP17} mechanism. This is widely used in machine learning \cite{SetTransformer,ParmarRVBLS19,VelikovicCCRLB2017,steenkiste2018relational,Mohart,Yang2019,ShapeContextNet} and based on the principle of keys, queries and values -- where each is a learned embedding of the input. The word `self' describes the fact that keys, queries and values are derived from the same context. In graph neural networks, this mechanism can be used to have nodes attend to their neighbours \cite{lin2017structured,VaswaniSPUJGKP17,VelikovicCCRLB2017,hoshen2017vain,shaw2018selfattention}. Each node serves as a focus point and queries information from the surrounding points. That is, the feature vector $f_i$ of node $i$ is transformed via an equivariant mapping into a query $q_i$. The feature vectors $f_j$ of the surrounding points $j$ are mapped to equivariant keys $k_{ij}$\footnote{Note that often, the keys and values do not depend on the query node, i.e. $k_{ij}=k_j$. However, in the SE(3)-Transformer, keys and values depend on the relative position between query $i$ and neighbour $j$ as well as on the feature vector $f_j$.}. A scalar product between key and query -- together with a softmax normalisation -- gives the attention weight. The scalar product of two rotation equivariant features of the same type gives an invariant feature. Multiplying the invariant weights $w_{ij}$ with the equivariant values $v_{ij}$ gives an equivariant output.
\section{Implementation of an Iterative SE(3)-Transformer}
Here, we describe the implementation of the iterative SE(3)-Transformer covering multiple aspects such as gradient flow, equivariance, weight sharing and avoidance of information bottlenecks.

\subsection{Gradient flow in Single-Pass vs. Iterative SE(3)-Transformers}

\label{sec:grad_flow}

\begin{figure}
    \centering
    \includegraphics[width=0.75\textwidth]{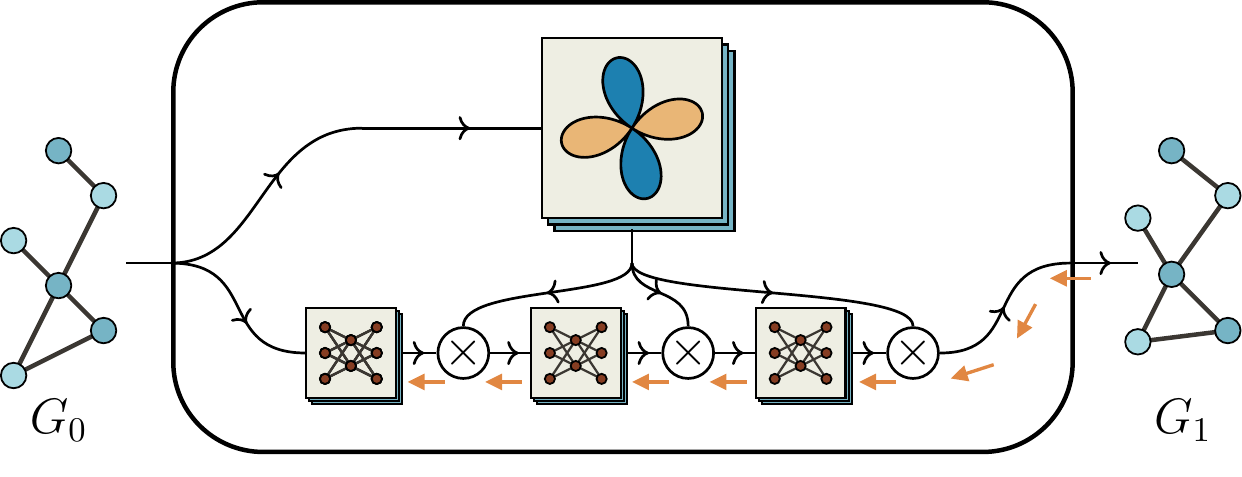}
    \caption{Gradient flow (orange) in a conventional, single-pass SE(3)-Transformer mapping from a graph (left) to an updated graph (right). The equivariant basis kernels (top) do not have to be differentiated as there is no gradient flow through them.}
    \label{fig:single_architecture}
\end{figure}

At the core of the SE(3)-Transformer are the kernel matrices $\mathbf{W}(\vec{x}_j - \vec{x}_i)$ which form the equivariant linear mappings used to obtain keys, queries and values in the attention mechanism. These matrices are a linear combination of basis matrices. The weights for this linear combination are the output of trainable neural networks. Importantly, the basis matrices are not learnable. They are defined by spherical harmonics and Clebsch-Gordan coefficients, and depend only on the relative positions in the input graph $G_0$\cite{SE3Transformer}.

Typically, equivariant networks have been applied to tasks where 3D coordinates in the input are mapped to an invariant or equivariant output in a single pass. This means that the relative positions of nodes do not change until the final output, and the basis matrices therefore remain constant. Gradients do not flow through them, and the spherical harmonics do not have to be differentiated, as can be seen in \Cref{fig:single_architecture}.

\begin{figure}
    \centering
    \includegraphics[width=0.75\textwidth]{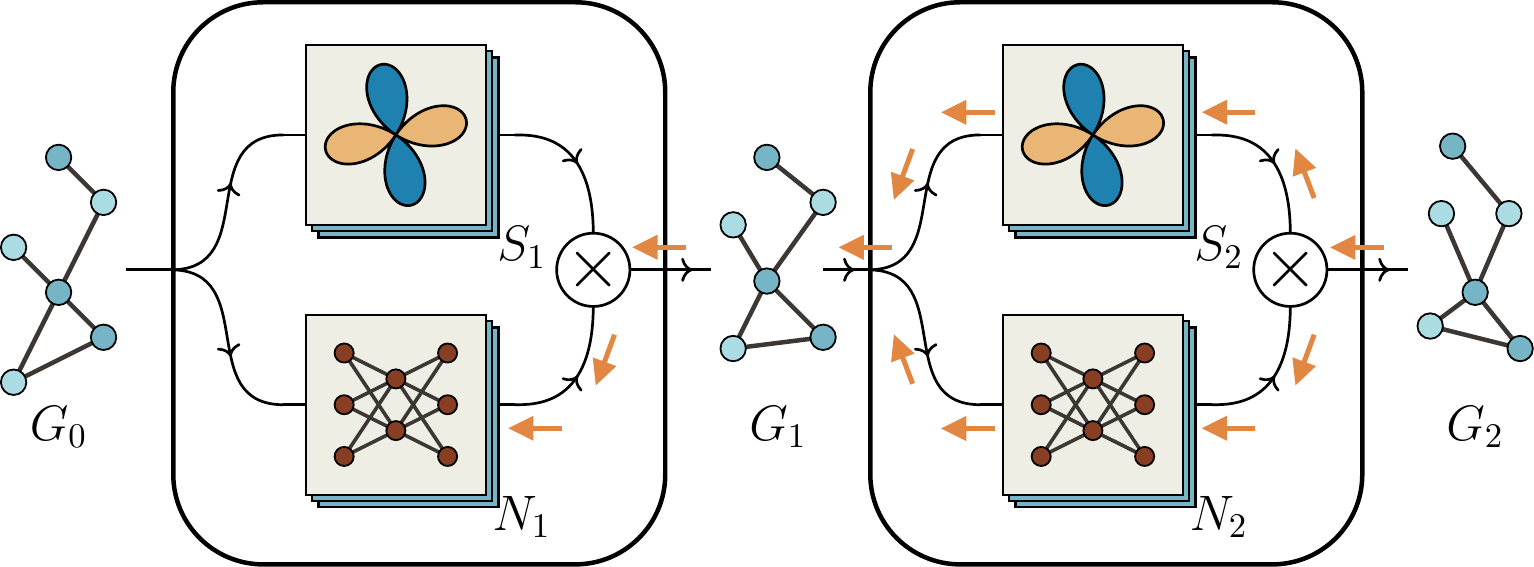}
    \caption{Gradient flow (orange) in an iterative SE(3)-Transformer with multiple position updates. Now the gradients do flow through the basis construction ($S_2$) meaning this part has to be differentiated.}
    \label{fig:iterated_architecture}
\end{figure}

When applying the SE(3)-Transformer in an iterative fashion (see \Cref{fig:iterated_architecture}), each block $i$ outputs an updated graph $G_i$. This allows for, e.g., re-evaluating the interactions or binary potentials between two nodes. 
Now the relative positions, and therefore the basis matrices, are no longer constant until the final output, and gradients flow through the basis.
The spherical harmonics used to construct the basis are smooth functions, and therefore backpropagating through them is possible. We provide code which implements this backpropagation.

\subsection{Hidden Representations Between Blocks}
\label{sec:hidden}
In the simplest case, each SE(3)-Transformer block outputs a single type-1 feature per point, which is then used as a relative update to the coordinates before applying the second block. This, however, introduces a bottleneck in the information flow. Instead, we choose to maintain the dimensionality of the hidden features. A typical choice would be to have the same number of channels (e.g. 4) for each feature type (e.g., up to type 3). In this case the hidden representation reads \texttt{\{0:4,1:4,2:4,3:4\}}. We then choose this same fiber structure for the outputs of each SE(3)-Transformer block (except the last) and the inputs to the blocks (except the first). This way, the amount of information saved in the hidden representations is constant throughout the entire network.

\subsection{Weight Sharing}
If each iteration is expected to solve the same sub-task, weight sharing can make sense in order to reduce overfitting. The effect on memory consumption and speed should, however, be negligible during training, as the basis functions still have to be evaluated and activations have to evaluated and stored separately for each iteration. In our implementation, we chose not to share weights between different SE(3)-Transformer blocks. The number of trainable parameters hence scales linearly with the number of iterations. In theory, this enables the network to leverage information gained in previous steps for deciding on the next update. One benefit of this choice is that it facilitates using larger fibers between the blocks as described in \ref{sec:hidden} to avoid information bottlenecks. A downside is that it fixes the number of iterations of the SE(3)-Transformer.

\subsection{Gradient Descent}
In this paper, we will apply the SE(3)-Transformer to an energy optimisation problem, loosely inspired by the protein structure prediction problem. For convex optimsation problems, gradient descent is a simple yet effective algorithm. However, long amino acid chains with a range of different interactions can be assumed to create many local minima. Our hypothesis is that the SE(3)-Transformer is better at escaping the local minimum of the starting configuration and likely to find a better global minimum. We add an optional post-processing step of gradient descent, which optimises the configuration within the minimum that the SE(3)-Transformer found. In real-world protein folding, these two steps do not have to use the same potential. Gradient descent needs a differentiable potential, whereas the SE(3)-Transformer is not subject to that constraint.


\section{Experiments}

\begin{figure}
    \centering
    \begin{subfigure}{0.44\textwidth}
        \centering
        \includegraphics[width=\textwidth]{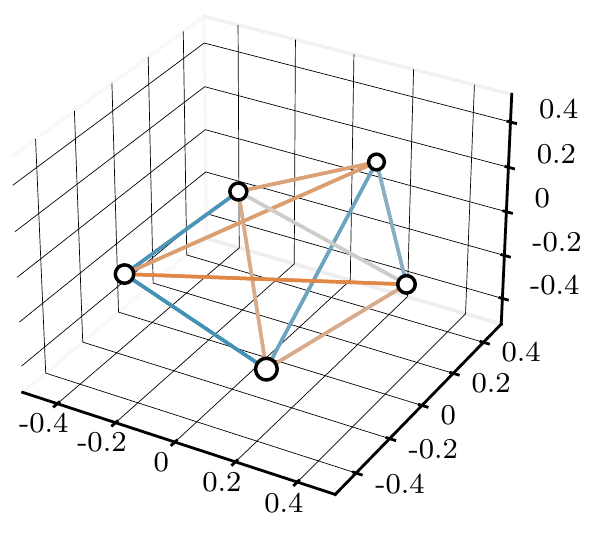}
    \end{subfigure}
    \quad\quad\quad
    \begin{subfigure}{0.35\textwidth}
        \centering
        \vspace{5mm}
        \includegraphics[width=\textwidth]{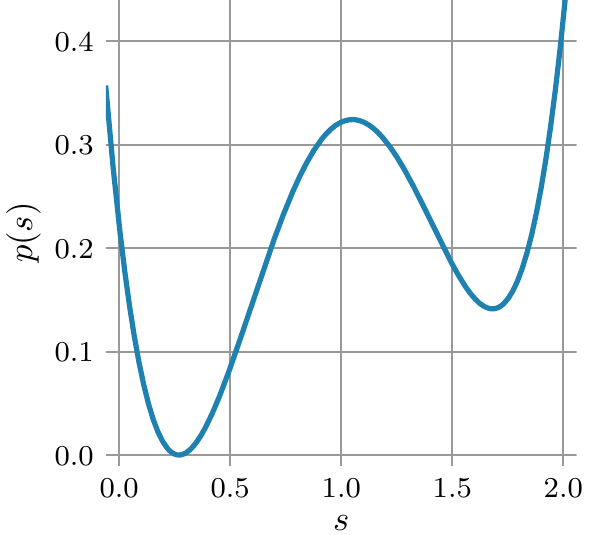}
    \end{subfigure}
    \caption{The left plot shows a configuration of five nodes, with the gradient of the potential between each pair represented by the colour of the edges. Blue edges indicate repulsion and orange edges indicate attraction. Stronger colour represents stronger interaction. The right plot shows the double-minimum potential $p(s)$, with parameter $a=0$.}
    \label{fig:experiment_cartoon}
\end{figure}

We study a physical toy problem as a proof-of-concept and to get insights into what type of tasks could benefit from iterative predictions. We consider an energy minimisation problem with 10 particles, which we will refer to as nodes. Each pair $(n_i, n_j)$ of nodes in the graph interacts according to a potential $p_{ij}(r_{ij})$, where $r_{ij}$ is the distance between the nodes. The goal is to minimise the total value of the potential across all pairs of nodes in the graph. 

We choose a pairwise potential with two local minima. This creates a complex global potential landscape that is not trivially solvable for gradient descent:
\begin{align}
s_{ij} &= r_{ij} - a_{ij} - 1 \\
p_{ij}(s_{ij}) &= s_{ij}^4 - s_{ij}^2 + \frac{s_{ij}}{10} + p_{\text{min}}
\end{align}
Here $p_{\text{min}} \approx 0.32$ ensures that $p_{ij}(s_{ij})$ attains a minimum value of zero. The parameter $a_{ij} = a_{ji}$ is a random number between $0.1$ and $1.0$ -- this stochasticity is necessary to avoid the existence of an optimal solution for all examples.

We consider three models for solving this problem: (i) the \textbf{single-pass} SE(3)-Transformer ($12$ layers); (ii) a three-block \textbf{iterative} SE(3)-Transformer ($4\times 3$ layers); (iii) \textbf{gradient descent} (GD) on the positions of the nodes. We also evaluate a combination of first applying an SE(3)-Transformer and then running GD on the output.
Additionally, we evaluate the iterative SE(3)-Transformer both with and without propagation of basis function gradients as described in \Cref{sec:grad_flow}.
We run GD until the update changes the potential by less than a fixed tolerance value. We train the SE(3)-Transformers for 100 epochs with 5000 examples per epoch, which takes about 90 minutes. 
We ran each model between 15 and 25 times -- the $\sigma$ values in Tables \ref{tab:results1} and \ref{tab:results2} represent $1\text{-}\sigma$ confidence intervals (computed using Student's t-distribution) for the mean performance achieved by each model. All models except GD have the same overall number of parameters.

\begin{table}[]
{\caption{Average energy of the system after optimisation (lower is better).}
\label{tab:results1}}
\vspace{1mm}
\resizebox{0.98\columnwidth}{!}{
\centering
\begin{tabular}{lrrrrr}
\toprule
& \hspace{4mm} \textbf{Gradient Descent} & \hspace{4mm} \textbf{Single-Pass} & \hspace{4mm}  \textbf{No Basis Gradients} & \hspace{4mm} \textbf{Iterative} & \hspace{4mm} \textbf{Iterative + GD} \\
\midrule
Energy & $0.0619$ & $0.0942$ & $0.0704$ & $0.0592$ & $0.0410$ \\
$\sigma$ & $\pm0.0001$ & $\pm0.0002$ & $\pm0.0025$ & $\pm0.0011$ & $\pm0.0003$ \\
\bottomrule
\end{tabular}
}
\end{table}

The results in \Cref{tab:results1} show that the iterative model performs significantly better than the single-pass model, approximately matching the performance of gradient descent.
The performance of the iterative model is significantly degraded if gradients are not propagated through the basis functions.
The best performing method was the SE(3)-Transformer followed by gradient descent.
The fact that the combination of SE(3)-Transformer and gradient descent outperforms pure gradient descent demonstrates that the SE(3)-Transformer finds a genuinely different solution to the one found by gradient descent, moving the problem into a better gradient descent basin.

\begin{table}[]
{\caption{Performance comparison of single-pass and iterative SE(3)-Transformers for different neighbourhood sizes $K$ and a fully connected version (FC).
\label{tab:results2}}}
\vspace{1mm}
\resizebox{0.98\columnwidth}{!}{
\centering
\begin{tabular}{lrrrrrr}
\toprule
& \hspace{4mm} \textbf{FC Single (K9)} & \hspace{4mm} \textbf{FC Iterative (K9)} & \hspace{4mm}  \textbf{K5 Single} & \hspace{4mm} \textbf{K5 Iterative} & \hspace{4mm} \textbf{K3 Single} & \hspace{4mm} \textbf{K3 Iterative} \\
\midrule
Energy & $0.0942$ & $0.0592$ & $0.1321$ & $0.0759$ & $0.1527$  & $0.0922$ \\
$\sigma$ & $\pm0.0002$ & $\pm0.0011$ & $\pm0.0003$ & $\pm0.0050$ & $\pm0.0001$ & $\pm0.0036$\\
\bottomrule
\end{tabular}
}
\end{table}

So far, every node attended to all $N-1$ other nodes in each layer, hence creating a fully connected graph. In \Cref{tab:results2}, we analyse how limited neighborhood sizes \cite{SE3Transformer} affect the performance, where a neighborhood size of $K=5$ means that every node attends to 5 other nodes. This reduces complexity from $\mathcal{O}(N^2)$ to $\mathcal{O}(NK)$, which is important in many practical applications with large graphs, such as proteins.
We choose the neighbors of each node by selecting the nodes with which it interacts most strongly. In the iterative SE(3)-Transformer, we update the neighbourhoods in each step as the interactions change.

\Cref{tab:results2} shows that the iterative version consistently outperforms the single-pass version across multiple neighborhood sizes, with the absolute difference being the biggest for the smallest $K$. In particular, it is worth noting that the iterative version with $K=3$ outperforms the fully connected single-pass network.

In summary, the iterative version consistently outperforms the single-pass version in finding low energy configurations. We emphasise that this experiment is no more than a proof-of-concept. However, we expect that when moving to larger graphs (e.g. proteins often have more than $10^3$ atoms), being able to explore different configurations iteratively will only become more important for building an effective optimiser.

\section*{Acknowledgements}
We thank Georgy Derevyanko, Oiwi Parker Jones and Rob Weston for helpful discussions and feedback. This research was funded by the EPSRC AIMS Centre for Doctoral Training at the University of Oxford and The Open Philanthropy Project Improving Protein Design.

\bibliographystyle{unsrt} 
\bibliography{references}

\newpage
\appendix
\section{Network Architecture and Training Details}
For the single-pass SE(3)-Transformer we use 12 layers. For the iterative version we use 3 iterations with 4 layers in each iteration. In both cases, for all hidden layers, we use type-0, type-1, and type-2 representations with 4 channels each. The attention uses a single head. The model was trained using an Adam optimizer with a cosine annealing learning rate decay starting at $10^{-3}$ and ending at $10^{-4}$. Our gradient descent implementation uses a step size of $0.02$ and stops optimizing when the norm of every position update is below $0.001$.

\end{document}